\documentclass{article}

\usepackage{arxiv}
\usepackage{amsmath}

\usepackage[utf8]{inputenc} 
\usepackage[T1]{fontenc}    
\usepackage{hyperref}       
\usepackage{url}            
\usepackage{booktabs}       
\usepackage{amsfonts}       
\usepackage{nicefrac}       
\usepackage{microtype}      
\usepackage{lipsum}
\usepackage{graphicx}
\graphicspath{ {./images/} }

\title{Teacher-Student Knowledge Distillation for Radar Perception on Embedded Accelerators}

\author{
Steven Shaw, \\
  Aptiv Technical Center\\
  Carmel, Indiana, 46032 \\
  \texttt{steven.shaw@aptiv.com} \\
   \And
Kanishka Tyagi,\\
  Aptiv Advance Research Center\\
  Agoura Hills, California, 91301\\
  \texttt{kanishka.tyagi@aptiv.com} \\
  \And
  Shan Zhang  \\
  Aptiv Advance Research Center\\
  Agoura Hills, California, 91301\\
  \texttt{shan.zhang@aptiv.com} \\
}

\begin{document}
\maketitle
\begin{abstract}
Many radar signal processing methodologies are being developed for critical road safety perception tasks. Unfortunately, these signal processing algorithms are often poorly suited to run on embedded hardware accelerators used in automobiles. Conversely, end-to-end machine learning (ML) approaches better exploit the performance gains brought by specialized accelerators. In this paper, we propose a teacher-student knowledge distillation approach for low-level radar perception tasks. We utilize a hybrid model for stationary object detection as a teacher to train an end-to-end ML student model. The student can efficiently harness embedded compute for real-time deployment. We demonstrate that the proposed student model runs at speeds 100x faster than the teacher model. 
\end{abstract}


\section{Introduction}
\label{sec:intro}
With the steady advances in autonomous driving, advanced safety features using one or more sensors are highly desirable. In order to avoid collisions and unintended breaking maneuvers, it is crucial to detect potential road obstacles accurately. Although camera and LiDAR-based object detection have been studied in the literature \cite{yahya2020object, qian20223d}, it’s only recently that interest in radar-based object detection using ML methods has begun, primarily because of its low cost, long-range detection capability, and robustness to poor weather conditions.

Traditionally, automotive radar-based object detection is performed through peak detection using simple local thresholding methods such as the Constant False-Alarm Rate (CFAR) algorithm \cite{richards2014fundamentals}. With the breakthroughs of ML in numerous applications \cite{zhang2017classification, zhang2019fusion, kumar2022material, TYAGI20223}, radar-based object perception using ML has attracted attention \cite{scheiner2020seeing, niederlohner2022self, akita2019object, major2019vehicle, zhang2020object, cozma2021deephybrid}. In \cite{scheiner2020seeing, niederlohner2022self}, point cloud radar data was used for object detection. Low-level radar spectra-based object perception was studied in  \cite{akita2019object, major2019vehicle, zhang2020object, cozma2021deephybrid}.

Radar data can be represented in many forms, such as point cloud, compressed data cube (CDC), or low-level spectra. Point cloud data is a detection level representation retaining the least amount of the original information. CDC \cite{mirko2021} consists of a subsection of beam vectors (BVs) that exceeds a CFAR threshold. Unlike point cloud and CDC, low-level spectra retains all of the information returned from the radar. In this work, we propose an ML based stationary object detection system using low-level spectra data. \cite{scheiner2020seeing, niederlohner2022self, akita2019object, major2019vehicle, zhang2020object, cozma2021deephybrid}, study large and mostly non-stationary objects such as pedestrians, cars, bikes, and trucks were considered for perception tasks. In contrast, our work focuses on detecting stationary debris objects in the driving path. 

Labeling low-level radar spectrum data is typically expensive. Large-scale deep models in \cite{mirko2021, akita2019object, major2019vehicle, zhang2020object, cozma2021deephybrid} often require a large amount of labeled data. With limited low-level radar spectrum data, a hybrid system can be used for perception tasks, e.g., object detection. However, many of the constituent traditional signal processing techniques cannot make use of hardware acceleration. Hybrid systems and large-scale deep models have computation efficiency requirements which make practical deployment to embedded systems a great challenge.
 
The essential factors for computing hardware deployed in next generation electric automobiles are cost and power efficiency. To achieve these goals, hardware accelerators such as matrix multiplication accelerators (MMA) are becoming increasingly popular. However, with MMA, deploying hybrid systems or even large-scale ML models on devices with limited resources, such as automotive, is still challenging. One solution to avoid performance loss is to take advantage of lightweight ML models that can efficiently run on MMA’s. Since we typically develop hybrid systems or large-scale deep models for perception tasks, knowledge distillation (KD) from these models (teacher models) to student models can enable fast inference on embedded systems \cite{wang2021knowledge}.
 
In this paper, we propose what we believe is the first teacher-student KD approach for stationary object detection using low-level radar spectrum data, where detection knowledge is transferred from a block-by-block (BB) hybrid teacher model to a light weight end-to-end ML student model intended for deployment to an embedded hardware accelerator. 

Overall, our main contributions are as follows. First, customized modules in the BB hybrid teacher model, i.e., the non-trainable blocks, effectively take advantage of the radar-specific properties. Second, the BB hybrid teacher model does automated labeling purely on low-level radar data and requires minimum to no human labeling efforts. Third, the proposed student model is a direct embedded implementation with low power and memory usage. Fourth, the student-teacher model works entirely on low-level radar datasets for stationary object detection problems. i.e., no other sensor is used to make the detection decisions; last, new evaluation methods for the teacher-student model is proposed and justified for its effectiveness. 

\section{Problem formulation}
\begin{figure}[ht]
\begin{center}
\includegraphics[height=3.5cm, width = 6.0cm]{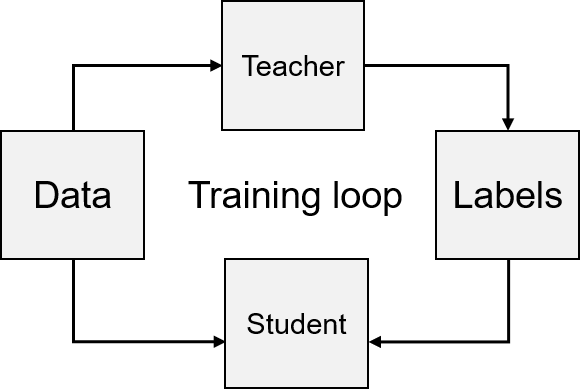}
\caption{Teacher-Student framework.}
\label{fig:teacher_train_loop}
\end{center}
\end{figure}

In our teacher-student framework shown in Figure \ref{fig:teacher_train_loop}, knowledge from a hybrid teacher is transferred to an end-to-end deep learning-based student. Since we are focusing on stationary object detection, the radar spectra can be reduced to a range azimuth spectrum maps. Therefore, the input spectrum data for the teacher and student models are range azimuth maps. The teacher model processes these range azimuth maps, creating labeled training data. The student model is then trained in a supervised fashion to mimic the same detection task that the teacher model performed. The student can later be easily deployed on an embedded platform and run much faster, more efficiently, and is product ready for the automotive market. Following \cite{hinton2015distilling}, \cite{wang2021knowledge}, we modify the idea of KD by changing the teacher model. In our work, the teacher is a BB hybrid model containing trainable and non-trainable parametric blocks.
By combining the traditional radar knowledge and deep learning approach in a BB hybrid model, we can significantly reduce the data needed for training \cite{gavrishchaka2018advantages} and the time-consuming data labeling efforts in prior end-to-end deep learning studies   \cite{akita2019object, major2019vehicle, cozma2021deephybrid}. 

\begin{figure}[ht]
\begin{center}
\centerline{\includegraphics[height=3.5cm, width=\columnwidth]{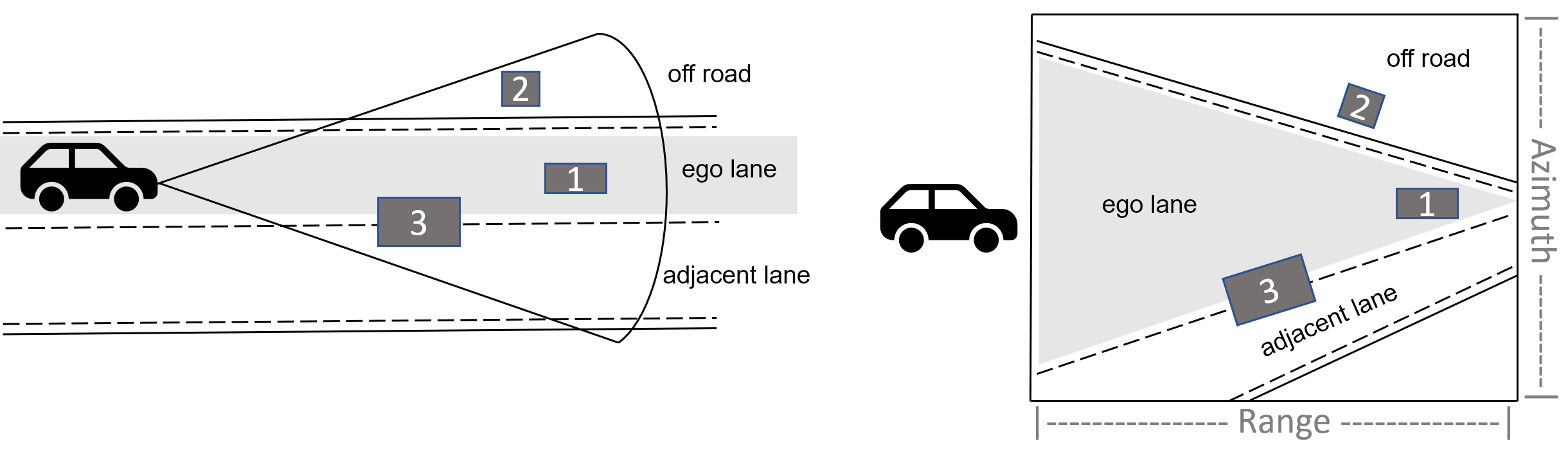}}
\caption{Cartesian and polar representation of debris detection for in-lane, out-lane objects.}
\label{fig:pol_cart}
\end{center}
\end{figure}
\vspace{-0.3in}

\section{Proposed Algorithm}
\begin{figure}[h]
    \centering
        \includegraphics[width=1.0\linewidth, height=0.4\textwidth]{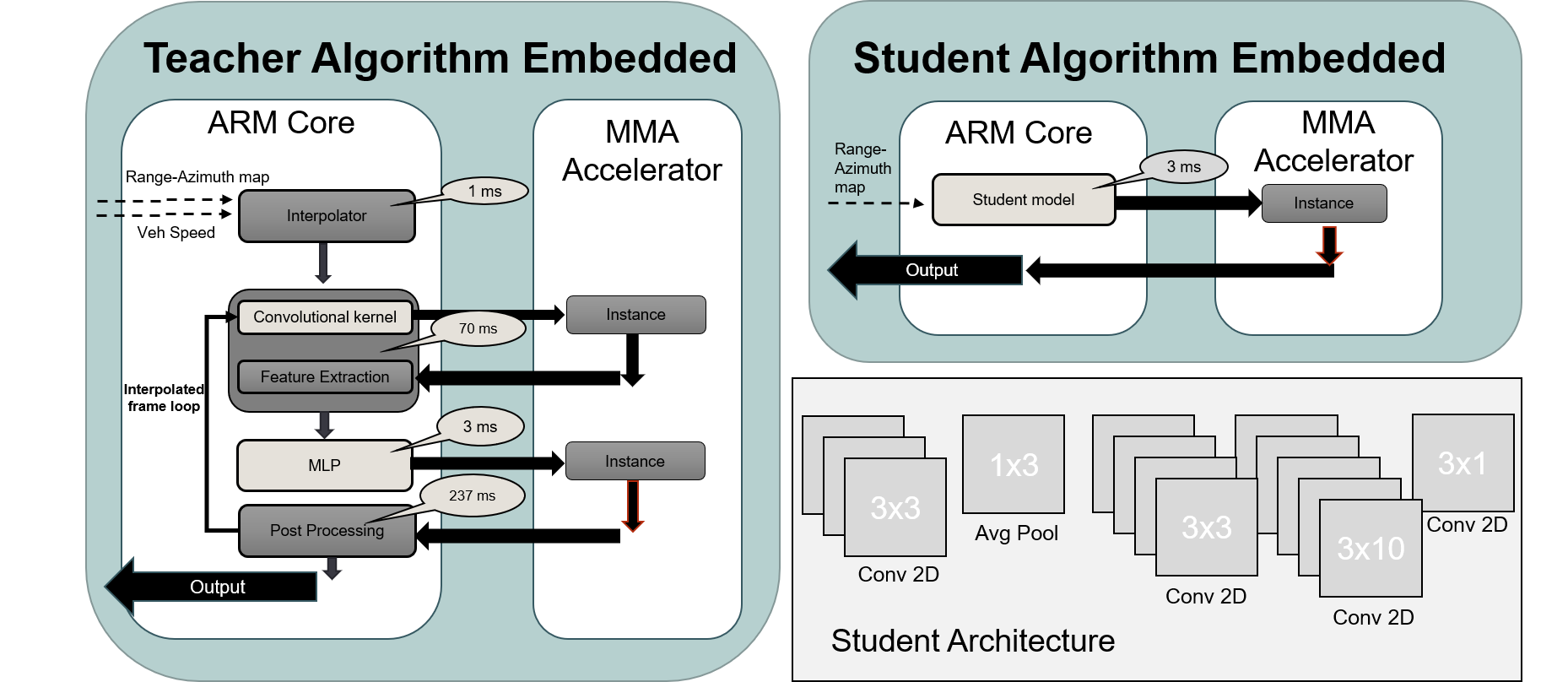}
        \caption{Teacher-Student embedded architecture. The timings for each block (in milliseconds) demonstrate the advantage of student algorithm. }
        \label{fig:bigfig}
\end{figure}

The teacher model contains sophisticated radar processing algorithms, some of which are trainable (e.g. a multi-layer perceptron (MLP)) and some non-trainable (e.g., interpolation, convolution based signal processing, feature extraction, and post processing). The non-trainable algorithms cannot benefit from the MMA core on an embedded devices, designed to run fast matrix multiplication calculations as in Figure \ref{fig:bigfig}. In our study, we observed that the teacher model's non-trainable components take the majority of the processing time, making it unsuitable to deploy on an embedded devices.

The teacher model takes as inputs a range azimuth map which is obtained by processing low-level time series radar data using traditional radar signal processing methods \cite{richards2014fundamentals} and host vehicle speed information. The teacher model uses the host vehicle speed for input spectrum interpolation. Data interpolation is needed for the purpose of manual feature extraction.  We assume that we have $N$ number of range azimuth samples. We use $\mathbf{X}_i \in \mathbb{R}^{464 \times 256}$ to denote the range azimuth map at time $i$, where $464$ is the number of range bins and $256$ is the number of azimuth bins, and $i = 1, \cdots, N$. The output of the teacher model is a probability vector containing probabilities of in-lane objects at each range bin over time. After applying a decision boundary on the probability vector, we obtain a in-lane object prediction vector $\mathbf{y}_i \in \mathbb{R}^{464 \times 1}$ at time $i, i = 1, \cdots, N$. We use $\mathbf{Y} \in \mathbb{R}^{N \times 464}$ denote the output decisions of the teacher model for all the data samples, where $\mathbf{Y}_{i, j} \in \{0, 1\}$ is the detection decision at range bin $j, j = 0, \cdots, 463$, and  $\mathbf{Y}_{i,j} = 1$ denotes that there is an obstacle at range bin $i$, otherwise, $\mathbf{Y}_{i,j} = 0$.

The teacher model contains radar processing tasks involving ego lane determination using specified geometry information, interpolation of the spectrum data, feature extraction with accumulated information across time, and a MLP detection network, to produce labels for the student model. The student model must learn all the mentioned tasks using only the input spectrum data and provided labels. Further more, the student model is able to provide high detection confidence without feature accumulation over time.

Although the raw input to the student model is the same shape as the teacher model, the first step taken by the student is to is to narrow its focus to the middle angular bins. We will denote the input range azimuth map for the student model by $\Tilde{\mathbf{X}}_i^s \in \mathbb{R}^{464 \times 30}, i = 1, \cdots, N$, where $30$ is the number of azimuth bins we select. These 30 angular bins ensure to capture of the widest portion of the ego lane, including a small buffer. Angular bin selection significantly improves the training time. 

Now, we present the network structure details of the student model. The student model comprises of five convolutional layers with ReLU activations and a Sigmoid activation for the final convolutional layer. Deploying fully connected layers on an embedded MMA accelerators is typically limited to such an extent as to not useful. Typically the solution is to utilize 1x1 convolution kernels to emulate fully connected layers. In our case, we use a 3x1 convolutional kernel in the final layer. Since objects usually span multiple range bins, using kernels with extended receptive field in the range direction allows the model to use the information from neighboring range bins to make a better predictions. Figure \ref{fig:bigfig} details the teacher-student paradigm along with the architectural details of each block. 

The collected radar dataset is extremely imbalanced. There are 464 range bins but typically just one in-lane object which may only span a few range bins. Therefore, we propose to use a weighted mean squared error (WMSE) to boost the loss values where in-lane objects are located by the teacher model, which is given as 
\vspace{-0.2in}
\begin{equation}
    \text{WMSE} = \frac{1}{N}\sum_{i=1}^{N}w_i(\hat{\mathbf{Y}}_{i, j}- \mathbf{Y}_{i, j})^2,
\label{eq:wmse}
\end{equation}
where $\hat{\mathbf{Y}} \in \mathbb{R}^{N \times 464}$ is the predicted probability score matrix, $\hat{\mathbf{Y}}_{i, j}$ is the predicted score and $\mathbf{Y}_{i, j}$ is the ground truth for time frame $i$ and range bin $j$.  
$w_i = 1$ when $\mathbf{Y}_{i, j} = 0$ and $w_i = \frac{|\mathbf{Y}^{-}|}{|\mathbf{Y}^{+}|}$ when $\mathbf{Y}_{i, j} = 1$, $i = 1, \cdots, N$ and $j = 0, \cdots, 463$.
 Also, $|\mathbf{Y}^{-}|$ denotes the number of negative samples and $|\mathbf{Y}^{+}|$ denotes the number of positive samples.

In the following, we show the training process of the student model. The input range azimuth maps are processed by the teacher algorithm to produce corresponding labels.
The labels are used to train the student model in a supervised setting. Knowledge distillation during training from the teacher to student model is illustrated in Figure \ref{fig:teacher_train_loop}. 

In order to discriminate between in- and out-of-lane objects, the student model must learn the ego lane geometry which is a function range as shown in Figure \ref{fig:pol_cart}. However, learning this function using solely convolutional layers presents a problem because convolutions are translationally invariant. To address this, we implement a modified version of the CoordConv algorithm \cite{cordconv} by appending two additional channels to label the range and azimuth respectively. This allows the convolution kernels to utilize their location on the range azimuth map in order to learn the lane width function. In our study, we found that CoordConv also allows the student to better model the apparent difference in signal strength between near and far objects.

For an input range-azimuth $\widetilde{\mathbf{X}}_{i}^{s} \in \mathbb{R}^{464 \times 30}, i = 1, \cdots, N$, we add two more range CoordConv and azimuth CoordConv channels, denoted by $\mathbf{R} \in \mathbb{R}^{464 \times 30}$ and $\mathbf{A} \in \mathbb{R}^{464 \times 30}$, respectively. The improved input data is denoted by $\mathbf{X}_i^s$ $\in$ $\mathbb{R}^{464 \times 30 \times 3}, i = 1, \cdots, N$. $\mathbf{X}_{i}^{s} =[\mathbf{x}_{i}^{s_{0}}\ \mathbf{x}_{i}^{s_{1}} \ \mathbf{x}_{i}^{s_{2}}]$ where $\mathbf{x}_{i}^{s_{0}} = \widetilde{\mathbf{X}}_{i}^{s}$, $\mathbf{x}_{i}^{s_{1}} = \mathbf{R}$, and $\mathbf{x}_{i}^{s_{2}} = \mathbf{A}$. $\mathbf{R}$ and $\mathbf{A}$ are given as:

\vspace{-0.1in}
\begin{align}
\mathbf{R}[n, :] =  \frac{n} {R_{\text{max}}},  \\
\mathbf{A}[:, m] =  2 |\frac{m - \frac{A_{\text{max}}}{2}} {A_{\text{max}}}|,\\
\end{align}
where $n, n = 0, \cdots, 463$ is the index for the range bin, $m, m = 0, \cdots, 29$ is the index for the azimuth bin, $\mathbf{R}_{\text{max}}$ is the maximum range index and $\mathbf{A}_{\text{max}}$ is the maximum azimuth index. $|\cdot|$ denotes absolute value operation.

The teacher model must look at multiple time steps before predicting the debris detection probability with reasonable confidence. Also, the teacher model interpolation step requires a minimum critical speed in order to make predictions. In order to avoid both of these issues, the student model is only trained with samples where the teacher was able to positively identify at least one debris object at some range. The intuition is to not penalize the student for extending the maximum distance at which debris can be detected as well as allow the student to fill in the gaps where the teacher could not make predictions due to slow host speed. The same intuitions underpinning the selective training technique have implications on the evaluation metric as discussed in the next section.

\section{Experimental Results}

Given the lack of a public low-level automotive radar dataset for early stationary object detection, we have collected our dataset. Our data collection vehicle has a high-definition radar mounted on the front bumper. Since it is costly to collect enough open environment data with stationary objects in-lane, the data we collected in a controlled environment is a set of debris objects including a dehumidifier, tire with and without rim, wooden pallet and stationary car next to repeated structures such as guardrails and signposts. The objects were placed 300m from the host vehicle before the beginning of each collection run. We also consider different rotations of each stationary object in order to get a complete characterization. The data was split into 70 \% training, 20 \% validation and 10 \% testing along with five fold validation. We used the Tensorflow library on a Windows i7 CPU to train the machine learning model. 

The setup of our experiment requires particular metrics for evaluating performance. We have modified the recall metric and call it R-score. R-score represents the percentage of true positives predicted by the student compared to the corresponding number of positives predicted by the teacher. Our R-Score can also allow a $+/- 1$ range bin offset allowance. We use $R_0$ to denote the score without range bin offset allowance, and $R_1$ to denote the score with  $+/- 1$ variance in the range bin direction. Note $R_0$ score is equivalent to recall, $\frac{tp}{tp + fn}$. $R_0$ and $R_1$ are given as
\vspace{-0.1in}
\begin{equation}
    R_0 = \frac{\sum_{i=1}^{N}\sum_{j=1}^{M} \mathbf{Y}_{i, j} \hat{\mathbf{Y}}_{i, j}^{'}}   {\sum_{i=1}^{N}\sum_{j=1}^{M}\mathbf{Y}_{i, j}},
\label{eq:Y0score}
\end{equation}

\begin{equation}
    R_1 = \frac{\sum_{i=1}^{N}\sum_{j=1}^{M} \mathbf{Y}_{i, j} \text{max}[\hat{\mathbf{Y}}_{i, j-1}^{'},\hat{\mathbf{Y}}_{i, j}^{'},\hat{\mathbf{Y}}_{i, j+1}^{'}] } {\sum_{i=1}^{N}\sum_{j=1}^{M}\mathbf{Y}_{i, j}},
\label{eq:Y1score}
\end{equation}
where $\mathbf{Y}$ is the ground truth and $\hat{\mathbf{Y}}^{'}$ is the predicted labels of the student model after a decision boundary is applied.

The student can extend the maximum range at which objects are detected. If all teacher samples were used as ground truth, the extended range performance of the student would be judged as false positives. Therefore, we have modified the precision score, $\frac{tp}{tp + fp}$, and call it P-score. $P_1$ precision score allows a $+/- 1$ range bin offset allowance for false positive identification. $P_0$ precision score is evaluated without any offset allowance. P-score and the true negative rate, specificity = $\frac{tn}{tn + fp}$, are only evaluated on samples which the teacher model has at least one positive prediction of an object at some range.

Table 1 shows results from the evaluation sets used in the experiment. The student extended the maximum range of debris detection by an average of 47 meters in the evaluation sets. The student scored an average of 0.91 for the $R_1$ recall score, an average of 0.93 for $P_1$ precision score, and an average of 0.99 for specificity score. The student also reduced the processing time on the desktop CPU benchmarks by an average factor 36 for the evaluation sets. This speed up is even greater on the embedded device where the student runs 100x faster than the teacher, see Figure \ref{fig:bigfig}.

\vspace{-0.1in}
\begin{table}[h]
\caption{Teacher (Tea) and Student (Stu) evaluation results comparison showing first detection range, average (Avg) runtime, statistical metrics. Range shown in meters and time in milliseconds. $R_0$, $R_1$, $P_0$, $P_1$, and sp denote $R_0$ recall, $R_1$ recall, $P_0$ precision, $P_1$ precision, and specificity, respectively.}
\begin{tabular}{|c|c|c|c|c|c|c|}
\hline
  & Bike & Tire & Shovel & Car & Toy Truck & Tire  \\ \hline
Stu max range      & 79  & 65   & 57  & 244 & 222 & 114 \\ \hline
Tea max range      & 8  & 5   & 26 & 215 & 168 & 76    \\\hline
Stu Avg Time    & 9  & 11  & 8  & 9 & 8 & 8    \\\hline
Tea Avg Time      & 323 & 318 & 329  & 299 & 311 & 325   \\\hline
$R_0$ & 0.83 & 0.88 & 0.90  & 0.98 & 1.00 & 0.82    \\\hline
$R_1$      & 0.9 & 0.88  & 1.00  & 1.00 & 1.00 & 0.82   \\\hline 
$P_0$ & 0.43 & 0.35 & 0.70  & 0.64 & 0.69 & 0.43    \\\hline
$P_1$      & 0.86 & 0.78  & 1.00  & 0.99 & 0.96 & 1.00   \\\hline 
$sp$      & 0.99 & 0.99  & 0.99  & 0.99 & 0.99 & 0.99   \\\hline 
\end{tabular}
\label{table:final_results}
\end{table}

\vspace{-0.1in}
\section{Conclusion}
The student model was successful in learning necessary mappings from the teacher. The student also exhibits many additional benefits over the teacher. The teacher's processing speed is dependent on the host vehicle's speed. As the host speed increases, the teacher produces more interpolated frames to processes. In contrast, the student processing time is independent of host speed allowing for constant runtime and memory usage, which are highly desirable attributes for embedded deployments. The student also exceeds the teacher's performance in the low-speed regime, where the teacher fails to make predictions. Through selective training, the student can take what it learns from the teacher's higher-speed examples and apply it to the low-speed samples. The student model vastly simplifies the deployment complexity onto embedded accelerators, all while running faster, using less memory and less power, and extending the maximum range of detections.

\bibliographystyle{unsrt}  
\bibliography{references}  






\end{document}